\documentclass{article}
\usepackage{spconf,amsmath,graphicx}

\usepackage[utf8]{inputenc}
\usepackage{spconf,graphicx}
\usepackage{comment}
\usepackage{amsmath,amssymb}
\usepackage{color}
\usepackage{url}
\usepackage[export]{adjustbox}
\usepackage{tcolorbox}

\usepackage{microtype}
\usepackage{subfigure}
\usepackage{booktabs}
\usepackage{multirow}
\usepackage{dsfont}
\usepackage{bbding}
\usepackage{pifont}
\usepackage{balance}
\usepackage{bm}
\usepackage{mathtools}
\usepackage{extarrows}
\usepackage{makecell}
\usepackage{arydshln}
\usepackage{multirow}
\usepackage{array}

\usepackage{bbm}
\usepackage{amssymb}
\usepackage{booktabs}

\usepackage{tikz}
\usepackage[edges]{forest} 
\usepackage{pgfplots}
\usepackage{scalefnt}
\usepackage[utf8]{inputenc}
\usepackage{arydshln} 
\usepackage{CJKutf8}

\title{SDIF-DA: A Shallow-to-Deep Interaction Framework with Data Augmentation for Multi-modal Intent Detection}

\name{Shijue Huang$^{1,2}$, Libo Qin$^{4*}$, Bingbing Wang$^{1,2}$, Geng Tu$^{1,2}$, Ruifeng Xu$^{1,2,3*}$\thanks{$^{*}$ Corresponding author. }
	\thanks{
		We thank the anonymous reviewers for their valuable suggestions to improve the quality of this work. This work was partially supported by the National Natural Science Foundation of China 62306342, 62176076, Natural Science Foundation of GuangDong 2023A1515012922, Shenzhen Foundational Research Funding JCYJ20220818102415032, Guangdong Provincial Key Laboratory of Novel Security Intelligence Technologies 2022B1212010005, The Major Key Project of PCL PCL2023A09.
		This work
		was also sponsored by CCF-Baidu Open Fund. 
}
}
\address{ $^1$School of Computer Science, Harbin Institute of Technology (Shenzhen)\\
	$^2$Guangdong Provincial Key Laboratory of Novel Security Intelligence
	Technologies\\
	$^3$Peng Cheng Laboratory~~ $^4$ School of Computer Science and Engineering, Central South University\\
	joehsj310@gmail.com,~~lbqin@csu.edu.cn,~~xuruifeng@hit.edu.cn
}

\begin{document}

\maketitle

\begin{abstract}
Multi-modal intent detection aims to utilize various modalities to understand the user’s intentions, which is essential for the deployment of dialogue systems in real-world scenarios. 
The two core challenges for multi-modal intent detection are (1) \textit{how to effectively align and fuse different features of modalities} and (2) \textit{the limited labeled multi-modal intent training data}. In this work, we introduce a shallow-to-deep interaction framework with data augmentation (SDIF-DA) to address the above challenges. Firstly, SDIF-DA leverages a shallow-to-deep interaction module to progressively and effectively align and fuse features across text, video, and audio modalities. Secondly, we propose a ChatGPT-based data augmentation approach to automatically augment sufficient training data. Experimental results demonstrate that SDIF-DA can effectively align and fuse multi-modal features by achieving state-of-the-art performance. In addition, extensive analyses show that the introduced data augmentation approach can successfully distill knowledge from the large language model.
\end{abstract}


\section{Introduction}
\label{Introduction}
\begin{figure}[t]
	\centering
	\includegraphics[width=0.37\textwidth]{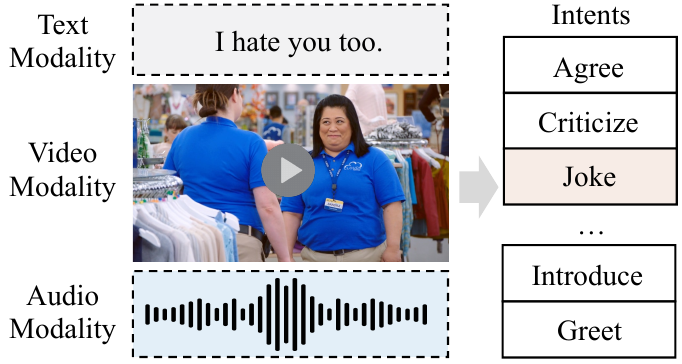}
		\vspace{-0.2cm}
	\caption{An example of multi-modal intent detection.}
		 \vspace{-0.4cm}
	\label{fig:intro}
\end{figure}

Intent detection is a core component of task-oriented dialogue systems, which is used to understand the current goal of users\cite{qin-etal-2019-stack,ijcai2021p622}.
Specifically, taking \figurename~\ref{fig:intro} as an example, solely based on the text information ``\textit{I hate you too.}'', we might infer that the intent of the speaker is to complain about someone incorrectly. In contrast, when we leverage the video and audio information, we can correctly observe that the intent of the speaker is joking, as both her expression and tone are exciting rather than angry. 

Recently, multi-modal intent detection has been developed to recognize the intention of users in multi-modal scenes, which is more practical in real-world scenarios and attracts increasing research attention.
To this end, a series of works focus on multi-modal intent detection including text and vision modalities~\cite{kruk-etal-2019-integrating,10.1145/3470482.3479636,maharana-etal-2022-multimodal}.
Another series of works study multi-modal intent detection consisting of text and audio modalities~\cite{9053281,9747759,dong22_interspeech}.
While promising progress has been witnessed, the above existing approaches mainly focus on the bi-modal setting, which is still insufficient for the real-world scenario that requires three modalities including text, audio, and vision modalities. To do this, 
Zhang \textit{et al.}~\cite{10.1145/3503161.3547906} introduce a public tri-modal intent detection benchmark, MIntRec, including text, video, and audio, to facilitate future research.
 However, such a scenario faces two major challenges: (1) \textit{how to effectively align and fuse features of different modalities} (i.e., text, video, and audio modalities) and (2) \textit{the lack of sufficient training data}.

To address the above challenges, we propose a \textbf{S}hallow-to-\textbf{D}eep \textbf{I}nteraction \textbf{F}ramework with \textbf{D}ata \textbf{A}ugmentation (SDIF-DA) to effectively fuse different modalities' features and alleviate the data scarcity problem.
Specifically, to solve the first challenge, SDIF-DA utilizes a shallow-to-deep interaction framework to fully achieve interaction across different modalities.
Firstly, it leverages a text-centric shallow interaction module to align the video and audio features to text modality in a shallow manner. Secondly, it introduces a transformer-based deep interaction module to further fuse all modalities' features in a deep fashion. 
To address the second challenge, we introduce a ChatGPT-based data augmentation approach to
 automatically augment sufficient training data, which has the advantage of naturally inferring knowledge from ChatGPT\footnote{https://openai.com/blog/chatgpt}. Finally, we obtain 25,000 augmented data and use them to improve the model training.

 \begin{figure*}[t]
	\centering
	\includegraphics[width=0.95\textwidth]{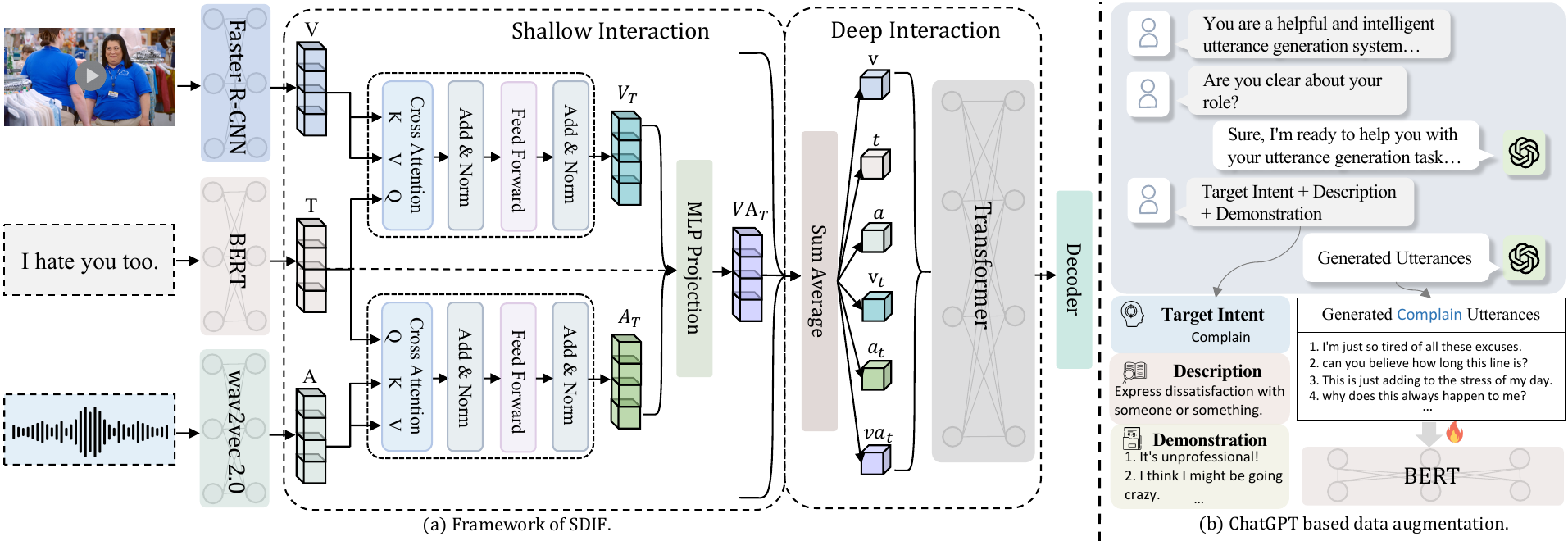}
		 \vspace{-0.3cm}
	\caption{(a) Overall architecture of shallow-to-deep interaction framework (SDIF). It contains a hierarchical module to conduct alignment in a shallow interaction manner, and a transformer module to aggregate and fuse all information with a deep interaction fashion; (b) Workflow of ChatGPT-based data augmentation approach.}
		\vspace{-0.3cm}
	\label{fig:SDIF}
\end{figure*}

 To summarize, our contributions are: (1) We propose a shallow-to-deep interaction framework for multi-modal intent detection, which can effectively and progressively align and fuse features of different modalities; (2) We introduce a ChatGPT-based data augmentation approach to automatically augment sufficient training data, which can alleviate the data scarcity problem in multi-modal intent detection; (3)  Experimental results and analyses on the existing benchmark show that our framework achieves the state-of-the-art performance and our data augmentation approach can distill useful knowledge from large language model. All codes are publicly available at \url{https://github.com/JoeYing1019/SDIF-DA}.

\section{Background}
This section describes the definition of multi-modal intent detection.
Specifically,
given a tri-modal input including text, video, and audio modalities, the multi-modal intent detection can be seen as a classification task to decide the intent label of tri-modal input, which is denoted as:
\begin{small}
	\begin{gather}
	y = \text{Model}(x_T, x_V, x_A), 
	\end{gather}
\end{small}where $x_T, x_V, x_A$ represents the text, video, and audio input, respectively; $y \in Y= \{y_1, y_2, ..., y_K\}$ is the corresponding intent label associated with one of the K existing intents. 

\section{Framework}
This section describes the architecture of shallow-to-deep interaction framework  (Section \ref{sec:SDIF}) and ChatGPT-based augmentation approach  (Section \ref{sec:da}), which are illustrated in \figurename~\ref{fig:SDIF}.
\subsection{Shallow-to-Deep Interaction Framework}\label{sec:SDIF}

\noindent \textbf{Feature Extraction.}
We follow Zhang et al.~\cite{10.1145/3503161.3547906} to extract the features of each modality. For text utterance $\bm{x}_T$, audio segment $\bm{x}_A$ and video segments $\bm{x}_V$, we adopt BERT~\cite{devlin-etal-2019-bert}, wav2vec 2.0~\cite{10.5555/3495724.3496768} and Faster R-CNN~\cite{NIPS2015_14bfa6bb} to conduct feature extraction, respectively:
\begin{small}
\begin{gather}
\bm{T} = \text{BERT}(\bm{x}_T),\\
\bm{A} =\text{Wav2Vec}(\bm{x}_A),\\
\bm{V} = \text{AvgPool}(\text{RoIAlign}(f,B)),
\end{gather}
\end{small}where the text feature $\bm T$ and audio feature $\bm A$ are obtained from encoder straightly; And for video modality, after extracting the feature representations $f$ of all keyframes by Faster R-CNN, the video feature $\bm V$ is obtained through mapping $f$ into the regions with the annotated bounding boxes $B$.

\noindent \textbf{Shallow Interaction.}
Previous studies have empirically shown that text modality is the most significant feature among multi-modal signals~\cite{10.1145/3462244.3479919, wu-etal-2021-text}. 
Inspired by this, we introduce a hierarchical module to align video and audio features with text feature to conduct shallow interaction as shown in \figurename~\ref{fig:SDIF}(a).

Specifically, the shallow interaction contains three layers:

(1) The first layer,
which comprises the unaligned origin representations, includes three extracted features: text modality $\bm T$, video modality $\bm V$, and audio modality $\bm A$.

(2) The second layer
adopts KQV attention~\cite{tsai-etal-2019-multimodal} to obtain bi-modal representations of video and audio aligned with text modality.
Concretely, we treat video/audio as source modality (obtaining key $K$ and value $V$) and text as target modality (obtaining query $Q$) to get video-text aligned feature $\bm{V_T}$ and audio-text aligned feature $\bm{A_T}$, which can be formulated as:
\begin{small}
\begin{gather}
	\bm{V_T} = \text{CrossAtt}(Q=\bm{T},K=\bm{V},V=\bm{V}), \\
	\bm{A_T} = \text{CrossAtt}(Q=\bm{T},K=\bm{A},V=\bm{A}).
\end{gather}
\end{small}
(3) The third layer, which 
includes a tri-modal representation that two other modalities simultaneously align with the text modality. 
To achieve this, we concatenate the text feature $\bm{T}$, the video-text aligned feature $\bm{V_T}$ and the  audio-text aligned feature $\bm{A_T}$, then project them into a new space by linear projection to get the tri-modal representation $\bm{\text{VA}_T}$:
\begin{small}
	\begin{gather}
	\bm{\text{VA}_T} = \text{MLP}(\bm{V_T} \oplus \bm{T} \oplus \bm{A_T}),
	\end{gather}
\end{small}where $\oplus$ is concatenation operation.

\noindent \textbf{Deep Interaction.}
 We apply transformer~\cite{10.5555/3295222.3295349} as our deep interaction module, which has been shown highly effective in performing modalities fusion~\cite{10.1145/3394171.3413678,10.1145/3380688.3380693,9897323}.
Before conducting the deep fusion, we first use the sum average method to fuse all features in shallow interaction module $H \in \{\bm{T},\bm{V}, \bm{A}, \bm{V_T}, \bm{A_T}, \bm{\text{VA}_T}\}$ into the corresponding overall representations $\bm h \in \{\bm{v}, \bm{t}, \bm{a}, \bm{v_t}, \bm{a_t}, \bm{\text{va}_t}\}$.
Then, we stack all the overall modality representations into a matrix  $M = [\bm{v}, \bm{t}, \bm{a}, \bm{v_t}, \bm{a_t}, \bm{\text{va}_t}]$. 
By setting $Q=K=V=M$, the transformer fuses all the representations through self-attention, which can be formulated as:
\begin{small}
	\begin{eqnarray}
	\bar{M}  = \text{Transformer}(M),
	\end{eqnarray}
\end{small}where $\bar{M} = [\bm{\bar{v}}, \bm{\bar{t}}, \bm{\bar{a}}, \bm{\bar{v_t}}, \bm{\bar{a_t}}, \bm{\bar{\text{va}_t}}]$ is the fused representations.

\noindent \textbf{Decoder.}
To make use of all complementary knowledge, we take the transformer output and construct a joint-vector $\bm{h^{out}}$ through concatenation. Then $\bm{h^{out}}$ is projected into label space using a linear decoder, which can be expressed as:
\begin{small}
	\begin{eqnarray}
	\bm {h^{out}} &=& \bm{\bar{v}} \oplus \bm{\bar{t}}  \oplus \bm{\bar{a}} \oplus \bm{\bar{v_t}} \oplus \bm{\bar{a_t}} \oplus \bm{\bar{\text{va}_t}}, \\
	\bm y &=& \text{MLP}(\bm {h^{out}}),
	\end{eqnarray}
\end{small}where $\bm{y}$ is the predicted output distribution.

\subsection{ChatGPT-based Data Augmentation Approach}\label{sec:da}
\noindent \textbf{Approach.}
Due to the lack of sufficient training data, we propose a data augmentation method based on ChatGPT, which has brought new opportunities and powerful abilities for generating text samples that resemble human-labeled data~\cite{zhou2023comprehensive, dai2023auggpt}.

As shown in \figurename~\ref{fig:SDIF} (b), 
we first utilize several turns of interaction to simulate the utterance generation ability of ChatGPT. 
Then to further describe the details of this task, we provide three types of information: 
(i) Target Intent: A intent like ``\textit{Complain}'' as the target of generated utterances; (ii) Description: The detailed description of the target intent such as ``\textit{Express dissatisfaction with someone or something}''; (iii) Demonstration: A certain number of utterances express target intent as demonstrations like ``\textit{It's unprofessional!}'', which helps ChatGPT further understand the meaning of target intent. 
We generate 1,250 utterances for each intent (20 taxonomies) and finally obtain 25,000 utterances, which is more than ten times the number of samples in MIntRec.

\noindent \textbf{Assist Learning.}
We use the augmented data to enhance the text feature extractor BERT, due to text modality playing a significant role in multi-modal fusion~\cite{10.1145/3462244.3479919, wu-etal-2021-text}. By treating the target intent and corresponding generated utterances as supervised data pairs, we use cross entropy for assist learning. 
\begin{small}
	\begin{gather}
	\mathcal{L}_{Aug}= -\bm {\hat{y_i}} \cdot log(\bm y_i) + (1-\bm {\hat{y_i}}) \cdot log(1-\bm y_i),
	\end{gather}
\end{small}where $\bm{\hat y_i}$ is gold label for sample $i$, $\bm y_i$ is the output predicted distribution for sample $i$.

\section{Experiment}

\begin{table}[t]
		\setlength\tabcolsep{3.5pt}
		 \renewcommand{\arraystretch}{1.15}
	\caption{\label{exp:main}  
		Main results on MIntRec. SDIF-DA is our shallow-to-deep interaction framework with data augmentation. 
		All baseline results are taken from Zhang \textit{et al.}~\cite{10.1145/3503161.3547906}.
		And \textit{w/o DA}, \textit{w/o SI}, and \textit{w/o DI} denote the removal of data augmentation, shallow interaction, and deep interaction, respectively.
     }
	\centering
		\begin{adjustbox}{width=0.49\textwidth}
	\setlength{\belowrulesep}{0 mm}{
		\setlength{\aboverulesep}{0 mm}{
			
	\begin{tabular}{l!{\vline}cccc!{\vline}cccc}
		\toprule
		\multirow{2}{*}{Methods}
		& \multicolumn{4}{c!{\vline}}{Twenty-class}& \multicolumn{4}{c}{Binary-class}\\
		
		& ACC(\%)  & F1(\%)  & P(\%)  & R(\%) & ACC(\%)  & F1(\%)  & P(\%) & R(\%) \\ 
		\midrule
		Text-Classifier & 70.88 & 67.40 & 68.07 & 67.44 & 88.09 & 87.96 &	87.95 &	88.09\\
		\midrule
		MAG-BERT~\cite{rahman-etal-2020-integrating} & 72.65 & 68.64 & 69.08 & 69.28  & 89.24 & 89.10 & 89.10  & 89.13 \\ 
		
		\midrule
		MulT~\cite{tsai-etal-2019-multimodal} & 72.52 & 69.25 & 70.25 & 69.24  & 89.19 & 89.07 & 89.02  & 89.18 \\
		
		\midrule
		MISA~\cite{10.1145/3394171.3413678} & 72.29 & 69.32 & 70.85 & 69.24  & 89.21 & 89.06 & 89.12  & 89.06 \\ 
		\midrule
		\textbf{SDIF-DA} & \textbf{73.71} & \textbf{71.58} & \textbf{72.43} & \textbf{71.21}  & \textbf{90.79} & \textbf{90.64} & \textbf{90.77}  & \textbf{90.53} \\ 
		\hdashline
		~~\textit{w/o DA }& 73.26 & 70.80 & 71.22 & 70.76 & 90.11 & 90.02& 89.91& 90.19 \\
		~~\textit{w/o SI}  & 68.54 & 64.90 & 65.09& 65.88 & 88.99  &  88.87 & 88.79 & 88.97 \\ 
		~~\textit{w/o DI } & 70.79& 67.61 & 67.57& 68.09 & 88.76 &  88.59 & 88.68 &88.51 \\
		\midrule
		\multirow{1}{*}{Human} & 85.51 & 85.07 & 86.37 & 85.74 & 94.72 & 94.67 & 94.74 & 94.82 \\
		\bottomrule 
	\end{tabular} }}
	\end{adjustbox}
 \vspace{-0.2cm}
\end{table}  

\begin{table*}[t]\small
	\centering  
	 \vspace{-0.2cm}
	\caption{Performance of different representation choices across three layers of shallow interaction module.}
	\label{exp:layer}
		\renewcommand{\arraystretch}{1.1}
	\begin{adjustbox}{width=0.7\textwidth}
			\setlength{\belowrulesep}{0 mm}{
			\setlength{\aboverulesep}{0 mm}{
		\begin{tabular}{@{\extracolsep{3pt}}ccc!{\vline}cc!{\vline}c!{\vline}cccc!{\vline}cccc}
			\toprule
			\multicolumn{3}{c!{\vline}}{Layer1}& \multicolumn{2}{c!{\vline}}{Layer2}& \multicolumn{1}{c!{\vline}}{Layer3}
			& \multicolumn{4}{c!{\vline}}{Twenty-class}& \multicolumn{4}{c}{Binary-class}\\
			\midrule
			V&T&A& $V_T$& $A_T$ & $\text{VA}_T$& ACC(\%)  & F1(\%)  & P(\%)  & R(\%) & ACC(\%)  & F1(\%)  & P(\%) & R(\%) \\ 
			\midrule
			$\checkmark$&$\checkmark$&$\checkmark$&$\times$&$\times$&$\times$&68.54&64.90&			65.09&65.88&88.99& 88.87 &	88.79&88.97 \\
			\midrule
			$\times$&$\times$&$\times$&$\checkmark$&$\checkmark$&$\times$&70.34&65.11&67.10&65.20&87.42& 87.09 &87.83&86.73 \\
			\midrule
			$\times$&$\times$&$\times$&$\times$&$\times$&$\checkmark$ &69.89&66.36&	68.28&			66.70&88.99& 88.92&	88.80 &89.23 \\
			\midrule
			$\times$&$\times$&$\times$&$\checkmark$&$\checkmark$&$\checkmark$&70.56&67.26&			67.17&68.22&89.89& 89.76&89.74&89.78 \\
			\midrule
			$\checkmark$&$\checkmark$&$\checkmark$&$\checkmark$&$\checkmark$&$\times$&71.91&67.61&	67.73&68.63&90.11& 89.98 &89.98&89.98	\\
			\midrule
			$\checkmark$&$\checkmark$&$\checkmark$&$\checkmark$&$\checkmark$&$\checkmark$& \textbf{73.71} & \textbf{71.58} & \textbf{72.43} & \textbf{71.21}  & \textbf{90.79} & \textbf{90.64} & \textbf{90.77}  & \textbf{90.53} \\ 
			\bottomrule 
		\end{tabular}}}
	\end{adjustbox}
	 \vspace{-0.3cm}
\end{table*}

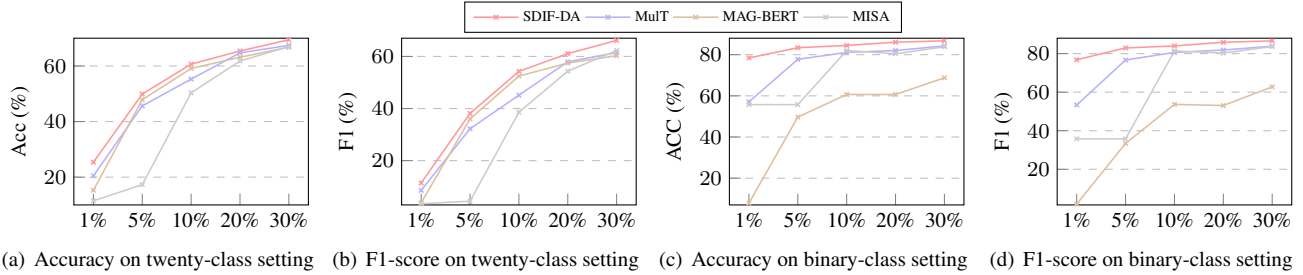
\begin{figure*}[t]
	\centering
	\subfigure[Accuracy on twenty-class setting]{
		\begin{minipage}[b]{0.23\textwidth}
			\centering
			\begin{tikzpicture}\footnotesize
				\begin{axis}[
					legend style={at={(0.5,-0.22)},
						anchor=north,legend columns=2},
					ylabel=Acc (\%),
					width=4.7cm, height=3.8cm,
								ymin=10, ymax=70,
					xtick={1, 2, 3, 4, 5},
					xticklabels={1\%, 5\%, 10\%, 20\%, 30\%},
					xlabel near ticks,
					ylabel near ticks,
					xtick pos=left,
					ytick pos=left,
					ytick align=inside,
					xtick align=inside,
					ymajorgrids=true,
					grid style=dashed,
					legend style={nodes={scale=0.6, transform shape}}, 
									every node near coord/.append style={font=\tiny},
					]
					\addplot+[semithick,mark=x,mark options={scale=0.6}, red!40!white] plot coordinates {
						(1, 25.39)
						(2, 49.89)
						(3, 60.67)
						(4, 65.39)
						(5, 69.44)
					};
					\addplot+[semithick,mark=x,mark options={scale=0.6}, blue!30!white] plot coordinates {
						(1, 20.45)
						(2, 45.62)
						(3, 55.28)
						(4, 64.72)
						(5, 67.42)
					};
					\addplot+[semithick,mark=x,mark options={scale=0.6}, brown!50!white] plot coordinates {
						(1, 15.28)
						(2, 47.87)
						(3, 59.10)
						(4, 63.15)
						(5, 66.74)
					};;
					\addplot+[semithick,mark=x,mark options={scale=0.6}, gray!40!white] plot coordinates {
						(1, 11.46)
						(2, 17.30)
						(3, 50.34)
						(4, 61.80)
						(5, 66.97)
					};
				\end{axis}
			\end{tikzpicture}
		\end{minipage}
		\label{fig:hor_2figs_1cap_2subcap_1}
	}
	\subfigure[F1-score on twenty-class setting]{
		\begin{minipage}[b]{0.23\textwidth}
			\centering
			\begin{tikzpicture}\footnotesize
				\begin{axis}[
					legend style={at={(1.2,1.22)},
						anchor=north,legend columns=4},
					ylabel=F1 (\%),
					width=4.7cm, height=3.8cm,
								ymin=3, ymax=67,
					xtick={1, 2, 3, 4, 5},
					xticklabels={1\%, 5\%, 10\%, 20\%, 30\%},
					xlabel near ticks,
					ylabel near ticks,
					xtick pos=left,
					ytick pos=left,
					ytick align=inside,
					xtick align=inside,
					ymajorgrids=true,
					grid style=dashed,
					legend style={nodes={scale=0.7, transform shape}}, 
									every node near coord/.append style={font=\tiny},
					]
					\addplot+[semithick,mark=x,mark options={scale=0.6}, red!40!white] plot coordinates {
						(1, 11.44)
						(2, 38.13)
						(3, 54.27)
						(4, 61.07)
						(5, 66.16)
					};
					\addplot+[semithick,mark=x,mark options={scale=0.6}, blue!30!white] plot coordinates {
						(1, 8.59)
						(2, 32.23)
						(3, 45.12)
						(4, 57.94)
						(5, 61.47)
					};
					\addplot+[semithick,mark=x,mark options={scale=0.6}, brown!50!white] plot coordinates {
						(1, 3.81)
						(2, 36.02)
						(3, 52.46)
						(4, 57.44)
						(5, 60.29)
					};;
					\addplot+[semithick,mark=x,mark options={scale=0.6}, gray!40!white] plot coordinates {
						(1, 3.45)
						(2, 4.43)
						(3, 38.59)
						(4, 54.33)
						(5, 62.30)
					};
					\legend{SDIF-DA, MulT, MAG-BERT, MISA}
				\end{axis}
			\end{tikzpicture}
		\end{minipage}
		\label{fig:hor_2figs_1cap_2subcap_2}
	}
	\subfigure[Accuracy on binary-class setting]{
		\begin{minipage}[b]{0.23\textwidth}
			\centering
			\begin{tikzpicture}\footnotesize
				\begin{axis}[
					legend style={at={(0.5,-0.22)},
						anchor=north,legend columns=2},
					ylabel=ACC (\%),
					width=4.7cm, height=3.8cm,
								ymin=7, ymax=88,
					xtick={1, 2, 3, 4, 5},
					xticklabels={1\%, 5\%, 10\%, 20\%, 30\%},
					xlabel near ticks,
					ylabel near ticks,
					xtick pos=left,
					ytick pos=left,
					ytick align=inside,
					xtick align=inside,
					ymajorgrids=true,
					grid style=dashed,
					legend style={nodes={scale=0.6, transform shape}}, 
									every node near coord/.append style={font=\tiny},
					]
					\addplot+[semithick,mark=x,mark options={scale=0.6}, red!40!white] plot coordinates {
						(1, 78.43)
						(2, 83.37)
						(3, 84.49)
						(4, 86.07)
						(5, 86.72)
					};
					\addplot+[semithick,mark=x,mark options={scale=0.6}, blue!30!white] plot coordinates {
						(1, 57.08)
						(2, 77.75)
						(3, 81.12)
						(4, 82.02)
						(5, 84.27)
					};
					\addplot+[semithick,mark=x,mark options={scale=0.6}, brown!50!white] plot coordinates {
						(1, 7.87)
						(2, 49.67)
						(3, 60.67)
						(4, 60.67)
						(5, 68.76)
					};;
					\addplot+[semithick,mark=x,mark options={scale=0.6}, gray!40!white] plot coordinates {
						(1, 55.73)
						(2, 55.73)
						(3, 82.02)
						(4, 80.45)
						(5, 83.82)
					};
				\end{axis}
			\end{tikzpicture}
		\end{minipage}
		\label{fig:hor_2figs_1cap_2subcap_3}
	}
	\subfigure[F1-score on binary-class setting]{
		\begin{minipage}[b]{0.23\textwidth}
			\centering
			\begin{tikzpicture}\footnotesize
				\begin{axis}[
					legend style={at={(0.5,-0.22)},
						anchor=north,legend columns=2},
					ylabel=F1 (\%),
					width=4.7cm, height=3.8cm,
								ymin=1.5, ymax=88,
					xtick={1, 2, 3, 4, 5},
					xticklabels={1\%, 5\%, 10\%, 20\%, 30\%},
					xlabel near ticks,
					ylabel near ticks,
					xtick pos=left,
					ytick pos=left,
					ytick align=inside,
					xtick align=inside,
					ymajorgrids=true,
					grid style=dashed,
					legend style={nodes={scale=0.6, transform shape}}, 
									every node near coord/.append style={font=\tiny},
					]
					\addplot+[semithick,mark=x,mark options={scale=0.6}, red!40!white] plot coordinates {
						(1, 76.83)
						(2, 82.94)
						(3, 83.98)
						(4, 85.90)
						(5, 86.61)
					};
					\addplot+[semithick,mark=x,mark options={scale=0.6}, blue!30!white] plot coordinates {
						(1, 53.34)
						(2, 76.67)
						(3, 80.63)
						(4, 81.85)
						(5, 83.86)
					};
					\addplot+[semithick,mark=x,mark options={scale=0.6}, brown!50!white] plot coordinates {
						(1, 1.95)
						(2, 33.44)
						(3, 53.67)
						(4, 53.07)
						(5, 62.76)
					};;
					\addplot+[semithick,mark=x,mark options={scale=0.6}, gray!40!white] plot coordinates {
						(1, 35.79)
						(2, 35.79)
						(3, 81.38)
						(4, 80.29)
						(5, 83.67)
					};
				\end{axis}
			\end{tikzpicture}
		\end{minipage}
		\label{fig:hor_2figs_1cap_2subcap_4}
	}
 \vspace{-0.2cm}
	\caption{Low-resource performance.}
	 \vspace{-0.3cm}
	\label{exp:low_resources}
\end{figure*}

\subsection{Experimental Settings}
We adopt \textit{bert-base-uncased}, \textit{wav2vec2-base-960h}, and Faster R-CNN as backbones of text, audio, and video feature extractor.
We use AdamW~\cite{loshchilov2018decoupled} as the optimizer.
The batch size and epoch for training are 8 and 100.
The learning rate is searched from $[5e^{-6}, 1e^{-5}]$.
We apply ChatGPT by OpenAI API and the corresponding version is \textit{gpt-3.5-turbo}. The number of given demonstration examples is set to 20.
All experiments are conducted at Tesla V100s.

\subsection{Main Results}
Following Zhang \textit{et al.}~\cite{10.1145/3503161.3547906}, we conduct experiments on the MIntRec with two settings, namely twenty-class and binary-class.
We adopt a series of state-of-the-art baselines including a text-only BERT-based appraoch \textit{Text-Classifier} and some tri-modal approaches: (1) \textit{MAG-BERT}~\cite{rahman-etal-2020-integrating}; (2)  \textit{MAG-BERT}~\cite{rahman-etal-2020-integrating}; (3) \textit{MISA}~\cite{10.1145/3394171.3413678}.
And we adopt four metrics, including accuracy (ACC), F1-score (F1), precision (P), and recall (R). 

The experimental results are shown in \tablename~\ref{exp:main}.
We can observe that
\textit{SDIF-DA} outperforms all the baselines in both the twenty-class setting and  binary-class setting with 1-3\% improvement on all metrics. This demonstrates that \textit{SDIF-DA} 
can effectively fuse different features across modalities.

\subsection{Analysis}
\noindent \textbf{All Components Contribute to the Final Performance.}
To further analyze whether all the components in \textit{SDIF-DA} make a contribution to the final performance, we conduct the ablation experiments including: removing \textit{shallow interaction (SI), deep interaction (DI), and data augmentation (DA) approach}. The results are illustrated in \tablename~\ref{exp:main}, we can observe that: (1) when removing the \textit{SI} or \textit{DI}, the performance both decline a lot, which indicates that the \textit{SI} and \textit{DI} are effective in aligning and fusing, respectively.
(2) When removing the \textit{DA}, the performance also declines but still outperforms all the baselines, this verifies the effectiveness of the data augmentation method.

\begin{figure}[t]
	\centering
	\begin{tikzpicture} [scale=0.9]\tiny
		\begin{axis}[
			enlargelimits=0.1,
			legend style={at={(0.5,-0.35)},
				anchor=north,legend columns=-1},
			ylabel={F1(\%)},
			ylabel near ticks,
			symbolic x coords={Classifier, MAG-BERT, MulT, MISA, SDIF, SDIF-DA},
			xtick=data,
			ybar=1.5pt,
			bar width=10pt,
			nodes near coords,
			nodes near coords align={vertical},
			nodes near coords style={font=\tiny},
			font=\footnotesize,
			grid=major,
			x tick label style={rotate=10},
			width=\linewidth,
			height=.45\linewidth,
			]
			
			\addplot [fill=brown!30!white,draw=brown]
			coordinates {
				(Classifier,32.32)  
				(MAG-BERT,33.97)
				(MulT,34.68)
				(MISA,36.15)
				(SDIF,44.44)
				(SDIF-DA,52.63)
			};
			\addplot [fill=red!15!white,draw=red]
			coordinates {
				(Classifier,46.12)  
				(MAG-BERT,47.09)
				(MulT,48.91)
				(MISA,46.44)
				(SDIF,55.56)
				(SDIF-DA,63.16)
			};
			
			\legend{Joke, Flaunt}
		\end{axis}
	\end{tikzpicture}
	\vspace{-0.2cm}
	\caption{Fine-grained analysis of two hard intent taxonomies.
		 SDIF denotes our framework without data augmentation.}
	\label{exp:fine-grained}
		\vspace{-0.4cm}
\end{figure}
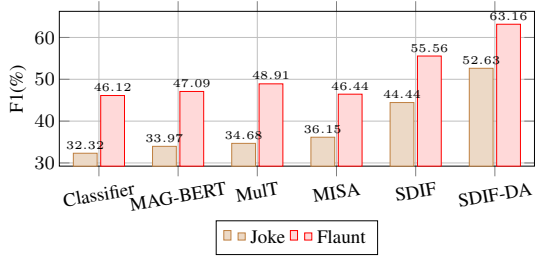

\noindent \textbf{All Layers in Shallow Interaction Make Effects.}
To explore whether all layers in shallow interaction make contributions. We conduct experiments by only selecting some subset of the three layers in shallow interaction.
The results are shown in \tablename~\ref{exp:layer}. 
We can observe that as the number of layers involved in the hierarchical architecture increases, the final performance shows an upward trend. Particularly, feeding all three layer representations achieves the best performance. 
This indicates that the features involved in each layer are complementary, and combining more of them results in better performance.

\noindent \textbf{\textit{SDIF-DA} Works in Low-resource Setting.}
To investigate the effectiveness of \textit{SDIF-DA} in low-resource scenarios and assess the impact of our data augmentation method, we conduct experiments using different limited training sizes, namely [{1\%, 5\%, 10\%, 20\%, 30\%}]. 
The results are shown in \figurename~\ref{exp:low_resources}, and we can observe that \textit{SDIF-DA} outperforms all baselines on all metrics, even when the available data resources are quite limited. This indicates that our approach can remain effective in low-resource scenarios and our ChatGPT-based data augmentation method can distill valuable knowledge from large language models, thus enhancing the performance of \textit{SDIF-DA} method, even with very limited training data.

\noindent \textbf{Fine-grained Analysis.}
To further explain why \textit{SDIF-DA} works, we analyze the F1-score of two fine-grained intent categories, namely, ``Flaunt'' (52 examples) and ``Joke'' (51 examples), which can be considered as the hard taxonomies that hold the almost most small number of samples in MIntRec. The experimental results are shown in \figurename~\ref{exp:fine-grained}. We can have two observations:
(1) \textit{SDIF captures fine-grained features}:
The \textit{SDIF} without data augmentation significantly surpasses all the baselines.
This indicates that our shallow-to-deep interaction framework can effectively capture fine-grained features, enabling accurate decision-making in challenging intent taxonomies.
(2) \textit{SDIF-DA distills knowledge from large language model}:
By incorporating the ChatGPT-based data augmentation approach, the performance of \textit{SDIF-DA} is further improved
 by a significant margin on both two categories.
 This suggests that our ChatGPT-based augmentation approach can successfully distill valuable knowledge from ChatGPT.

\section{Conclusion}
\label{conclusion}
In this work, we introduce a shallow-to-deep interaction framework with data augmentation (SDIF-DA)  for multi-modal intent detection.
The proposed shallow-to-deep interaction framework can successfully align and fuse different features across text, video, and audio modalities. In addition, the introduced ChatGPT-based data augmentation approach can automatically augment sufficient utterances to alleviate the data scarcity problem. Experimental results on existing benchmark show that SDIF-DA attains the state-of-the-art performance.
\bibliographystyle{IEEEbib}
\bibliography{strings,refs,custom}

\end{document}